\documentclass[lettersize,journal]{IEEEtran}
\usepackage{amsmath,amsfonts}
\usepackage{algpseudocode}
\usepackage{algorithm}
\usepackage{array}
\usepackage[caption=false,font=normalsize,labelfont=sf,textfont=sf]{subfig}
\usepackage{textcomp}
\usepackage{stfloats}
\usepackage{url}
\usepackage{verbatim}
\usepackage{graphicx}
\usepackage{color}
\usepackage{gensymb}
\def\BibTeX{{\rm B\kern-.05em{\sc i\kern-.025em b}\kern-.08em
    T\kern-.1667em\lower.7ex\hbox{E}\kern-.125emX}}
\usepackage{balance}
\begin{document}

\title{Robust Adversarial Attacks Detection based on Explainable Deep Reinforcement Learning for UAV Guidance and Planning}
\author{Thomas Hickling,~
        Nabil Aouf,~
        Phillippa Spencer
\thanks{ T. Hickling and N. Aouf are with the School of Mathematics, Computer Science and Engineering 
Department of Electrical and Electronic Engineering
City, University of London Northampton Square, London, EC1V 0HB. E-mails: \texttt{\{tom.hickling, nabil.aouf\}@city.ac.uk}.}
\thanks{P. Spencer is with the Defence Science and Technology Laboratory (DSTL), Porton Down, Salisbury, United Kingdom}}

\markboth{Transactions on Intelligent Vehicles}
{How to Use the IEEEtran \LaTeX \ Templates}

\maketitle

\begin{abstract}
The dangers of adversarial attacks on Uncrewed Aerial Vehicle (UAV) agents operating in public are increasing. Adopting AI-based techniques and, more specifically, Deep Learning (DL) approaches to control and guide these UAVs can be beneficial in terms of performance but can add concerns regarding the safety of those techniques and their vulnerability against adversarial attacks. Confusion in the agent's decision-making process caused by these attacks can seriously affect the safety of the UAV. This paper proposes an innovative approach based on the explainability of DL methods to build an efficient detector that will protect these DL schemes and the UAVs adopting them from attacks. The agent adopts a Deep Reinforcement Learning (DRL) scheme for guidance and planning. The agent is trained with a Deep Deterministic Policy Gradient (DDPG) with Prioritised Experience Replay (PER) DRL scheme that utilises Artificial Potential Field (APF) to improve training times and obstacle avoidance performance. A simulated environment for UAV explainable DRL-based planning and guidance, including obstacles and adversarial attacks, is built. The adversarial attacks are generated by the Basic Iterative Method (BIM) algorithm and reduced obstacle course completion rates from 97\% to 35\%. Two adversarial attack detectors are proposed to counter this reduction. The first one is a Convolutional Neural Network Adversarial Detector (CNN-AD), which achieves accuracy in the detection of 80\%. The second detector utilises a Long Short Term Memory (LSTM) network. It reaches an accuracy of 91\% with faster computing times compared to the CNN-AD, allowing for real-time adversarial detection.
\end{abstract}


\section{Introduction}
\IEEEPARstart{T}{he} ability to autonomously pilot UAVs may lead to many new use opportunities in the coming years. According to a Goldman Sachs report\cite{website:goldman}, it is expected that the UAV market, driven by commercial, defence and consumer needs, will reach a total of \$100 billion in the future. As UAVs become more widely used, the adoption of Artificial Intelligence (AI) has also increased with better hardware and software, leading to improved processing speed and better-performing algorithms. Combining these two technologies has resulted in autonomously guided UAVs, which achieve performance comparable to human operators, if not better \cite{app9245571}. 

One area of AI that is of particular interest in this field is Deep Reinforcement Learning (DRL). This scheme utilises reinforcement learning techniques and applies them to a Deep Learning (DL) framework. DRL uses Markov Decision Processes where a state leads to an action that produces a reward. These DRL schemes perform well in tasks with unstructured data as it learns using trial and error. The DRL approach applies in a variety of fields such as AI in computer games\cite{DRL-game}, robot control\cite{DRL-robot}, and natural language processing\cite{Wang}. This paper will adopt a DRL scheme to perform UAV guidance. DRL schemes have been used with UAVs before, such as in Liu et al.(2022)\cite{9916069}, where it was used to optimise the positioning of UAVs for data collection. The guidance task in this paper will use the Deep Deterministic Policy Gradient (DDPG) method with a Prioritised Experience Replay (PER). The DDPG scheme is a well-established algorithm, so this paper will look at increasing the efficiency in training and safety in operation by utilising an Artificial Potential Field (APF). This method produces a repulsive field around obstacles and an attractive field around the goal. Adding the APF will likely lead to faster training times and safer operation during flight.
 
Though DL can be a helpful tool, it has limitations in its explainability due to its closed-box nature. The opaqueness of DNNs can lead to a lack of trust in the decisions made by the agent. Research into the eXplainable Artificial Intelligence (XAI) has produced methods, such as SHapley Additive exPlanations (SHAP)\cite{lundberg2017unified}, Local Interpretable Model-agnostic Explanations (LIME)\cite{ribeiro2016should}, Decision Trees\cite{lundberg2020local}, that provide explanations to understand decisions made by the DL agents. Although these methods can lead to good explanations of the decisions made, there needs to be more work done in using these explanations to improve the performance of AIs. This paper aims to use SHAP to attempt to solve one of the significant problems in DL.

With the closed-box nature of these DL algorithms, unintended paths for making decisions arise. When an outside user exploits these paths, it is known as an adversarial attack\cite{Carlini}\cite{FGSM}. Adversarial attacks are created by making small perturbations in the data inputs that have statistically more significant impacts on the decision of the AI agents. These attacks are dependent on the known parameters of the attacked AI agent (closed-box attack or white-box attack) when the attack occurs (training or operation), and the methods of generating the attack (Fast Gradient Sign Method (FGSM)\cite{FGSM}, Projected Gradient Descent (PGD)\cite{madry2017towards}, Carlini and Wagner (C\&W)\cite{carlini2017provably}). This paper aims to use the FSGM attack to show a reduction in the successful operation of an AI-guided UAV.


This paper aims to develop a robust adversarial attack-based explainability detection scheme which suits the drone's DRL-based guidance technology during in-flight operation. The training of the UAV uses a neural network to perform a guidance task using the DDPG-PER with an APF DRL scheme. The SHAP method provides explanations of the agent's actions by giving those decisions context.

An attack is then used on the drone using an adversarial attack algorithm based on the FSGM designed to move the drone off course. Having insight into how the drone's planning and guidance decisions are made through explainability, detecting an altered decision from an attack is possible. We propose to monitor these slight variations in the decision process with deep network-based detectors. We introduce the first one based on a Convolutional Neural Network (CNN). Using the SHAP values generated by analysing the DDPG-APF agent's depth image input layer, this CNN-based Adversarial Detector (CNN-AD) learns and finds features in the image generated by the SHAP values to detect adversarial attacks. 

The second type of detector we propose is a Recurrent Neural Network (RNN) in the form of a Long Short Term Memory (LSTM) network. This detector monitors over a 10-time step period the changes in the generated SHAP values to determine if an attack occurs. The SHAP values are generated on a hidden layer within the DDPG-APF agent to reduce the number of SHAP values generated and considered. The hidden layer for the SHAP generation is the penultimate layer before the image and positional processes are combined. It is vital that the adversarial attack detector can perform real-time detection for the safe operation of the UAV. So a good detector can decide on the state before the next time-step.

The successful operation of autonomous drones will depend upon AI models' ability to work to a high degree of reliability and resiliency. Adversarial attacks could erode such systems' mission capability by manipulating the sensor data gathered by the UAV. By manipulating object avoidance systems or threat detection systems, adversarial attacks could deny capability and access to areas. These systems must be operated with limited oversight, as constant supervision will remove many benefits of autonomous drones. Therefore, an adversarial detection system should work with high certainty and in real time. This paper aims to provide a system that meets these criteria.\\

The main contributions of this paper are as follows:
{\color{black}
\\
\begin{itemize}
    \item Proposing and developing a method for adversarial attack generation upon a DRL-controlled drone for the first time in the literature.
    \item Proposing two innovative deep network-based methods exploiting explainability concepts to detect adversarial attacks upon DRL-based autonomous drone guidance.
    \item Showing the ability for real-time detection of adversarial attacks using one of the deep network-based detection methods proposed.
    \item Showing the proposed detection's effectiveness and favourable comparison with other alternatives.
    
\end{itemize}}
  
\section{Related Work}\label{sec2}

This paper builds and integrates elements of different domains (AI, UAVs, Planning and Guidance, and Adversarial Detection) to create an adversarial attack detector that uses a DRL guidance scheme adopted for drones. Here the most relevant papers to our own will be highlighted. 

Many papers are adopting DRL methods to build a guidance agent for a drone to complete an obstacle course safely. In Sang-Yun et al.(2019)\cite{app9245571}, they tested the ability of multiple DRL obstacle avoidance agents against differing skill level human operators. The paper finds that some DRL schemes can complete the most challenging obstacle courses while the human operators and the other algorithms do not reach the goal. Guo et al.(2020)\cite{guo2020autonomous} show that the extension of the APF to a DDPG scheme to control a ship can positively impact the training efficiency of the DDPG.

Showing how SHAP can be used to explain the decisions of a UAV in an obstacle course, He et al.(2021)\cite{HE2021107052} adopt a DRL algorithm (Twin Delayed DDPG (TD3)) for drone guidance. They combine this scheme with SHAP values to provide explainability to the TD3 agent. Using a second explainability method known as a Gradient-weighted Class Activation Mapping (Grad-CAM) to create a saliency map to show important regions in the input depth image. Using the SHAP values along with Grad-CAM reasonably explains the drone's actions. 

{\color{black}The papers that cover the adversarial attacks used in this paper are by Goodfellow et al.(2014)\cite{goodfellow2014explaining}, who first described the FGSM, and by Kurakin et al.(2017), who expanded on the FGSM by introducing the BIM. This reduces the distortion of the attacked images by reducing the $\varepsilon$ value and then iterating the process to achieve higher attack success rates at lower visual perturbation.}

To judge the adversarial detectors to be produced, looking at the detectors designed in other papers is valuable. Ilahi et al.(2022)\cite{9536399} review the different types of adversarial attacks on DRL agents and the methods to detect or nullify them. Looking at the papers in the survey that looked at adversarial detection, Lin et al.(2017)\cite{lin2017detecting} use an action-conditioned frame prediction module to detect attacks. The prediction module allows them to detect attacks and use predicted frames to make robust decisions, as they can lose any altered frames. This method can detect adversarial attacks with an accuracy of 60-100\% depending on the type of attack used.

Havens et al.(2018)\cite{havens2018online} use an agent that detects adversarial examples by the agent learning separate sub-policies using the Meta Learned Advantage Hierarchy (MLAH) framework. MLAH is a method to protect against adversarial attacks during the training period. The agent finds the attacks because they are unexpected due to the sub-policies. 

Xiang et al.(2018)\cite{xiang2018pca} set out a robot pathfinding task to test their adversarial detector. They designed their detector to consider five factors, energy point gravitation, key point gravitation, path gravitation, included angle, and the placid point. Using the data points, they calculate the probability of an adversarial attack on the robot. This method gained a precision of around 70\% detection. 

Though not using a DRL scheme in Fidel et al.(2020)\cite{fidel2020explainability}, the authors use the explanations generated by SHAP values to detect adversarial attacks upon an image classifier. The SHAP values generate a pattern that is unidentifiable as a particular class. Thus any pattern that falls outside this class can be detected as an attack. Their attack detection model consists of a fully connected network of four layers and produces detection rates ranging from 12.6\% to 100\% depending on the dataset and the type of adversarial attack considered. Most attacks are detected at a rate of over 90\%, with only a few specific cases below that. 

As this paper looks to use an LSTM network to detect adversarial attacks in a UAV guidance system, researchers have used LSTM networks to detect Distributed Denial of Service (DDoS)\cite{9288358} attacks on networks. These LSTM detectors have shown to be more robust than other network types because of the network's discrete property and the input samples' utility requirement. 

This paper attempts to build an autonomous UAV DRL-based guidance scheme. This guidance scheme will be robust against adversarial attacks by building a novel adversarial detector exploiting either CNN or LSTM network architectures that utilise the SHAP-based explainability of the DRL guidance scheme to detect attacks.

\section{Methodology}\label{sec3}
This section will lay out the main pillars of this paper: First, a short description of the DRL agent and the introduction of the APF technique to improve the performance of the DRL training efficiency and operational safety. Second, a brief outline of the generation of SHAP values as a method of XAI. Third, describe the methods to generate adversarial attacks upon the drone. Finally, explain the two novel methods for creating a DNN adversarial detection scheme.

\subsection{Deep Deterministic Policy Gradient with Prioritised Experience Replay and Artificial Potential Field}
A DRL method is chosen to control the drone's guidance. A move away from Q-learning has occurred for tasks in continuous action spaces. Their poor performance means that most DRL guidance models use Actor-Critic methods for training. There are a few leading examples of these methods, such as TD3\cite{fujimoto2018addressing}, TRPO\cite{schulman2015trust}, SAC\cite{haarnoja2018soft}, and DDPG\cite{lillicrap2015continuous}—this final one, DDPG that will be implemented as the DRL scheme for this study.

DDPG uses an actor-critic method that combines the policy gradient and the value function. It comprises two separate networks, the actor that controls the policy function and the critic that controls the value function. The construction of these networks is shown in Fig. 1.

The performance of the DDPG is improved further by being paired with a PER. The PER\cite{8122622} is a method for choosing higher relevancy states for training the model. These can be "states" that lead to higher or negative rewards, as the model can learn better from more impactful experiences. The PER has a probabilistic element so that the same "states" are not chosen too frequently.

This algorithm is already well-known by the team, so implementation is trivial. The DDPG is the only type of DRL scheme to be used as this paper is interested in something other than the scheme's performance relative to other DRL schemes. Comparisons between DRL schemes have been covered in other studies\cite{bouhamed2020autonomous}. 

\begin{figure*}
    \centering
    \includegraphics[width=14cm]{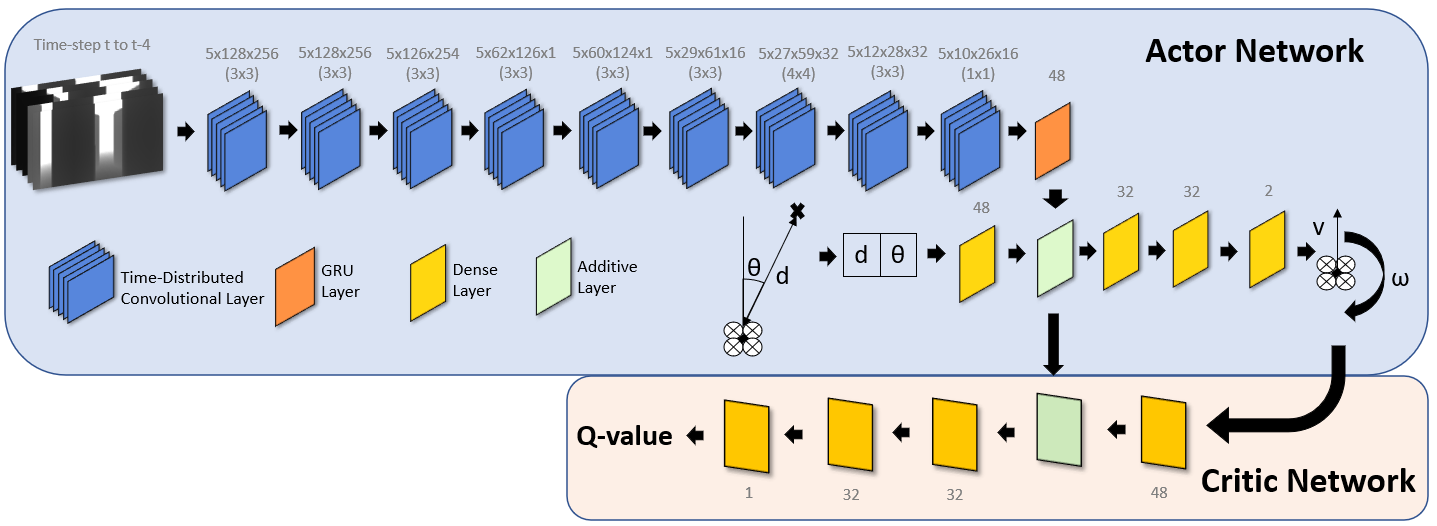}
    \caption{The construction of the two neural networks of the DDPG architecture. The actor-network controls the drone's movement, and the critic network generates the Q-value that trains the networks.}
    \label{fig:Actor Critic}
\end{figure*}

{\color{black}
\begin{equation}
F(v_x) = 
\left\{
    \begin{array}{lr}
         \frac{20\cos{\theta}}{d^2}, & \text{if } d<5 \\
         \frac{2\cos{\theta}}{d}, & \text{if } d\geq 5
    \end{array}
\right.
\end{equation}

\begin{equation}
F(\omega) = 
\left\{
    \begin{array}{ll}
         \tan{\frac{2\theta}{5}}, & \text{if } d<5 \\
         \tan{(\frac{\theta}{3}})^3, & \text{if } d\geq 5, \text{ and } |\theta| > \frac{\pi}{2}
    \end{array}
\right.
\end{equation}}

\begin{algorithm}[htp]
{\color{black}
\caption{DDPG with PER and APF}
\begin{algorithmic}

\State Initialise action-value network $Q(s_t,a_t,w)$, actor-network $\mu(s_t,\nu)$ with $w$ and $v$ separately, replay buffer B with size S;
\State Initialise target network $Q'(s_t,a_t,w)$, $\mu'(s_t,\nu)$ with $w$ and $v$ separately;
\State Initialise maximum priority, parameters $\alpha, \beta$, updating rate of the target network $\lambda$, minibatch K;
\For{episode = 1, M}
    \State Initialise a random noise process $\Phi$;
    \State Observe initial state $s_1$;
    \For{t=1, H}
        \State Calculate action:
        \State $a_t=a^*_t+\Phi+F_a(v_z,\omega)+\sum_iF_r(v_z,\omega)_i$ eq(1,2,3);
        \State Obtain the next state: $s_{t+1}=H(a_t)$;
        \State Obtain reward: $r_t=R(s_t,s_{t+1})$ eq(11);
        \State Store experience $(s_t,a_t,r_t,s_{t+1})$ in replay buffer \State B and set $D_t=max_{i<t}D_i$;
        \If{t>S}
        \For{j=1, K}
        \State Sample experience $j$ with probability $P(j)$:
        \State $P(j) = \frac{D_j^\alpha}{\sum_kD_k^\alpha}$;
        \State Compute corresponding importance \State sampling weight $W_j$ and TD-error $\delta_j$:
        \State $W_j=\frac{1}{S^\beta\cdot P(j)^\beta}$;
        \State $\delta _j=r(s_t,a_t)+\gamma Q'-Q$
        \State Update the priority of transition j \State according to absolute TD-error $|\delta_j|$;
        \EndFor
        \State Minimise the loss function to update \State action-value network: $L=\frac{1}{K}\sum_iw_i\delta_i^2$;
        \State Compute policy gradient to update the actor \State network: \State $\nabla_v\mu|_{s_i}\approx\frac{1}{K}\sum_i\Delta_aQ(s_t,a_t,w)\nabla_v\mu(s_t,v)$;
        \State Adjust the parameters of the target network\State $Q'(s_t,a_t,w)$ and $\mu'(s_t,v)$ with an \State update rate $\lambda$;
        \EndIf
    \EndFor
\EndFor
\end{algorithmic}}
\end{algorithm}

Here $\theta$ and $d$ are the bearings and the distance between the drone and the goal, respectively. $v_x$ is the forward speed while $\omega$ is the yaw rate. The force that acts upon the forward velocity of the drone is inversely proportional to the distance $d$. The angular force is proportional to the size of the bearing. There is an increase in the rate of force change at 5m. The angular force only applies when the absolute bearing is over 90\degree when further than 5m from the goal.  
For the repelling force from the obstacles, the functions are as follows:
{\color{black}
\begin{equation}
    F(\omega) = 
    \left\{
        \begin{array}{ll}
            (-3+3\tan{\frac{\theta_{ob}}{2}}), & \text{if } d_{ob}<3, \text{ \& } 0<\theta_{ob}<\frac{\pi}{2} \\
            (3+3\tan{\frac{\theta_{ob}}{2}}), & \text{if } d_{ob}<3, \text{ \&} -\frac{\pi}{2}<\theta_{ob}<0 \\
            0, & \text{if } d_{ob}\geq 3
        \end{array}
    \right.
\end{equation}}
The values $\theta_{ob}$ and $d_{ob}$ are the bearing and the distance between the drone and the obstacle, respectively. The range of the repellent field is limited to 5m, which is enough to reduce collisions. The repellent field only affects the yaw rate as this prevents collisions successfully. Fig. 3 shows how the two fields act upon the drone and the orientations. Fig. 2 shows graphically the training process with Algorithm 1 showing the pseudo-code.

\begin{figure}[htp]
    \centering
    \includegraphics[width=8.4cm]{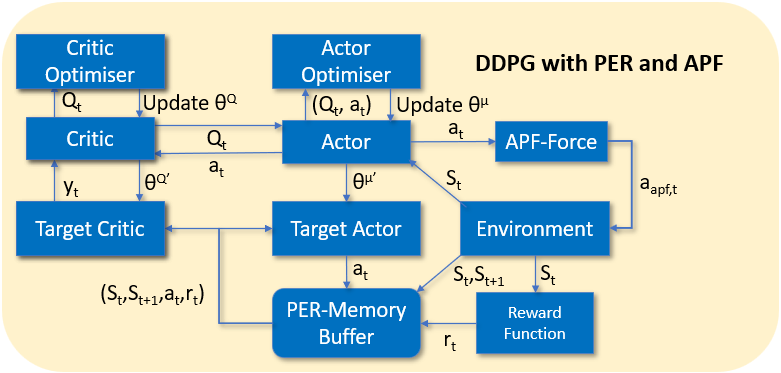}
    \caption{The flow of information during training to the various parts of the DDPG with PER and APF.}
    \label{fig:DDPG layout}
\end{figure}
\begin{figure}[htp]
    \centering
    \includegraphics[width=8.4cm]{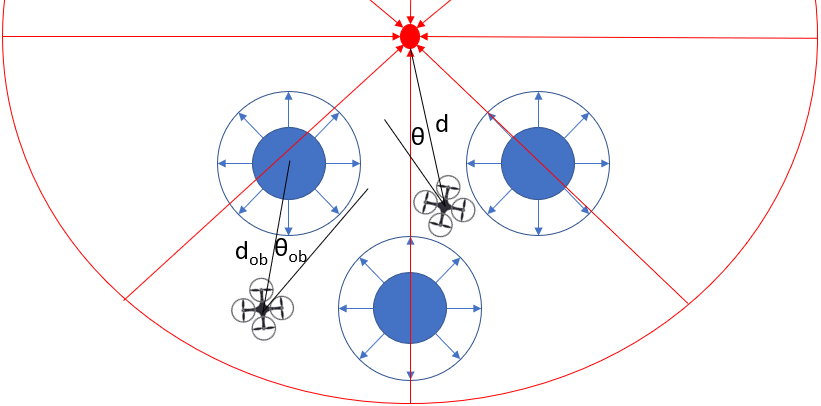}
    \caption{Diagram showing the attractive force in red and the repulsive force in blue. The two coordinate systems are shown between the goal, the drone, and the closest obstacle.}
    \label{fig:APF diagram}
\end{figure}

\begin{figure}[htp]
    \centering
    \includegraphics[width=8.4cm]{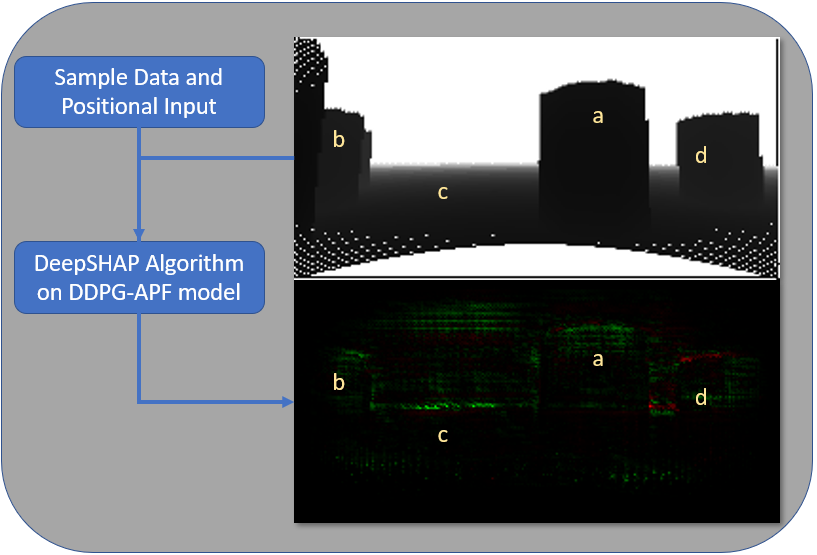}
    \caption{The SHAP value images are created using the DeepSHAP algorithm on the model with a sample set and the inputs to be analysed. The structures can be seen at (a), (b), and (d). The horizon is seen at (c).}
    \label{fig:SHAP images}
\end{figure}
\begin{figure*}[htp]
    \centering
    \includegraphics[width=12cm]{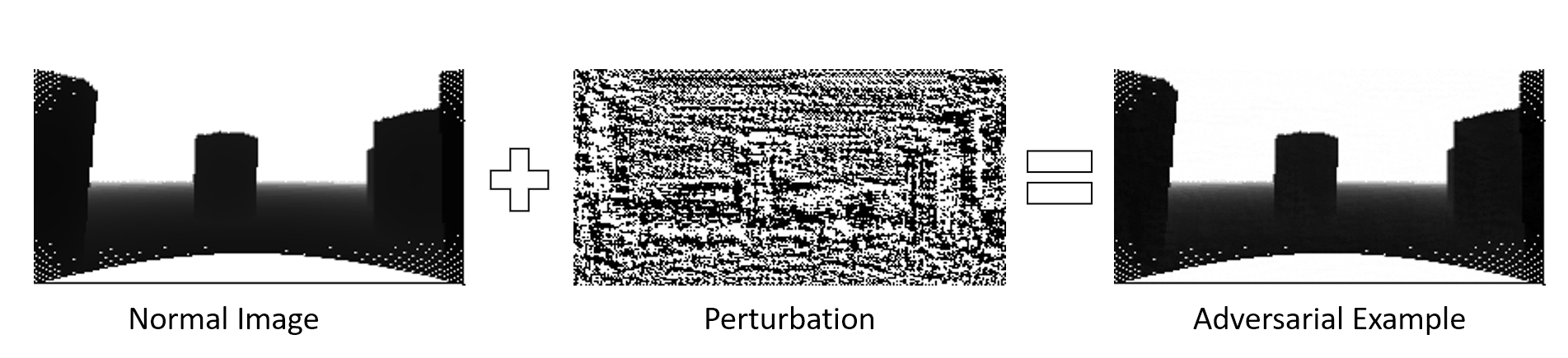}
    \caption{This figure shows how the perturbation is added to the original image, with the resultant image not noticeably changed. The perturbation is small, so this image has been exaggerated to show the structure.}
    \label{fig:FGSM example}
\end{figure*}
\subsection{Explainability Through Shapley Values}
AI decisions can be challenging to understand due to their closed-box nature. This lack of explainability leads to a lack of trust in AI models. The lack of understanding is especially true with non-experts and has slowed the adoption of AI technologies.

XAI methodologies split into making the model interpretable from the start or using post-hoc explanations for closed-box models. Post-hoc explanations come in the form of global or local explanations. Global explanations explain the average behaviour of the model, while local explanations explain individual predictions. Explanations come in the form of values indicating the importance of inputs, saliency maps for visual inputs, rule sets for creating decision trees, or generating natural language responses. 

The XAI method we need here must be post-hoc and local to generate explanations at every layer that the adversarial detector can analyse. Two methods considered are LIME\cite{ribeiro2016should} and SHAP. SHAP is preferred as it provides mathematical guarantees for the accuracy and consistency of explanations.

SHAP is built on a game theory concept known as Shapley values, where players are assessed on an individual contribution to a collective game. Values are assigned on the impact of each player on that final score. Traditional Shapley values cannot handle the extensive inputs used in DL, so approximation methods are produced. The SHAP framework is proposed by Lundberg and Lee\cite{lundberg2017unified} to fulfil this need.

SHAP works by approximating the Shapley values that assign a value to each feature based on the importance of the particular prediction. Shapley values are found using the following formula:
\begin{equation}
    \Phi_i=\sum\limits_{S\subseteq F\backslash{i}}\frac{|S|!(|F|-|S|-1)!}{|F|!}[f_{S\cup\{i\}}(x_{S\cup\{i\}})-f_S(x_S)]
\end{equation}
In this formula, $F$ is all the features while $S$ is a subset of the features. $f_{S\cup\{i\}}$ is a model that has been trained with the feature $i$ present and $f_S$ is the model trained without that feature. These two are then compared against each other with this section of the function $f_{S\cup\{i\}}(x_{S\cup\{i\}})-f_S(x_S)$ where $x_S$ is the value of the input feature. This value is computed for all subsets $S\subseteq F\backslash{i}$ because the change in feature depends on the other features in the model.

{\color{black}
Calculating the Shapley values can be computationally expensive for networks with many inputs. Thus, a method to approximate the Shapley values was devised by implementing DeepLIFT\cite{shrikumar2017learning} to create DeepSHAP, a framework for layer-wise propagation of Shapley values. DeepLIFT was designed to calculate the contributions of the inputs to the outputs by back-propagating the values through the neural network. Each layer's local Shapley values can be summarised in a linear equation using a reference value to get the delta values. Using these delta values and utilising the chain rule to progress through the network nodes until all contributions are known. The SHAP value calculation is shown in Equation 5; the SHAP value is the contribution of all the local SHAP values using the chain rule:

\begin{equation}
    \phi_i(f_i,y) \approx \Delta y \sum_i^n m_{y_if_j}m_{x_jf_i}
\end{equation}

The DeepSHAP method uses the rescale rule, and the reveal cancel rule to linearise the non-linear parts of a neural network. The rescale rule is used for functions like ReLU and Sigmoid. It splits contributions into positive and negative values leading to the multiplier used to be the ratio between $\Delta x$ and $\Delta y$. The reveal cancel rule is used for more complicated situations by splitting the ratio used for the rescale rule into negative and positive values. The rescale rule multiplier is given in Equation 6:

\begin{equation}
    m_{\Delta x^+ \Delta y^+}=m_{\Delta x^- \Delta y^-}=m_{\Delta x \Delta y}=\frac{\Delta y}{\Delta x}
\end{equation}
The multiplier for the reveal cancel rule is given by:
\begin{equation}
\begin{split}
    m_{\Delta x^+ \Delta y^+}=\frac{\Delta y^+}{\Delta x^+} \\
    m_{\Delta x^- \Delta y^-}=\frac{\Delta y^-}{\Delta x^-}
\end{split}
\end{equation}}

Two sets of values are produced when generating the SHAP values for a decision made by the DDPG-APF agent. The first set produces the contributions to the x-velocity, either positive for forward or negative for backward. The second set is for the yaw rate, either positive for clockwise (right) or negative for anti-clockwise (left). The SHAP values for a given depth image can be seen in Fig. 4.


\subsection{Adversarial Detection}
The complexity of DL models allows for unintended pathways to decisions in the action space that would not usually be taken. Researchers have exploited these pathways using adversarial attacks. These adversarial attacks create small perturbations in the DL model's input that alter their decisions. The negligible visual impact of the perturbation can be seen in Fig. 5.

As described in the previous section, XAI techniques such as SHAP explain the decisions of DNNs. Thus, it is possible to detect the use of different pathways and adversarial attacks with specially crafted neural networks. {\color{black}The proposed scheme's benefit is building an adversarial detector that can operate in real time and at a high detection rate. Most research in making UAVs safe from adversarial attacks has focused on making the model more robust, either by adversarial training, defensive distillation, or robust learning\cite{9536399}. These methods can protect from adversarial attacks, but they will not alert the user to the attack. Some previous methods for adversarial detection, such as frame prediction models, have been shown to work in simple atari models but may struggle with unconfined simulations due to the complexity of frame prediction, this also leads to significant delays in adversarial detection. Other methods have poor accuracy, which this method hopes to improve upon. One other benefit is the original model needs no alteration to accommodate this adversarial detector, meaning this solution becomes easy to implement. The proposed adversarial detectors will aim to be high-accuracy real-time detectors with the ability to plug and play into any DRL scheme. This will also be the first detection method shown to work in a UAV guidance application.} This section will discuss the method used to create the adversarial attacks and the two novel methods of DL to detect the changes in the SHAP values to uncover the attacks.

\subsubsection{Fast Gradient Sign Method and Basic Iterative Method}

A method to generate adversarial attacks is required based on the ease of implementation and the computation time. FGSM, developed by Goodfellow et al.(2014), \cite{goodfellow2014explaining} can generate adversarial examples quickly and is simple to implement. This method is also flexible as it allows the BIM technique to be applied to add iterations to strengthen the adversarial attack without just increasing $\varepsilon$.

{\color{black}
FGSM uses a gradient technique to generate adversarial attacks. The method uses the gradient of the loss function $\nabla_xL(\theta,x,y)$ to maximise the loss. The $\theta$ signifies the model parameters, while the $x$ is the model output, and the $y$ is the desired output. If this function is positive, the increase of $x$ will increase the loss function. A negative gradient function will increase the loss function by decreasing $x$. The adverse function $\Psi$ is shown below:
\begin{equation}
\begin{split}
    \Psi : X \times Y &\longrightarrow X \\
    (x,y) &\longrightarrow  \varepsilon.sign(\nabla_xL(\theta,x,y))
\end{split}
\end{equation}
From this a mask is created that will maximise the loss for the input to the model of $(x,y)$. The $\varepsilon$ value will decide how strong the attack is; an attack at maximum strength would have an $\varepsilon$ value of 1 which would produce an image that looks like random static to a human observer. An attack strength of 1/255 is a typical strength for an attack that is visually undetectable, this number is usually chosen as it is the minimum size of a pixel value in an 8-bit colour channel. The $\varepsilon$ value can be varied to get the effect that is required depending on attack strength and image quality. This mask of strength of $\varepsilon$ is then combined with the sample $x$ to produce $x'$:}
\begin{equation}
    x'=x+\varepsilon.sign(\nabla_xL(\theta,x,y))
\end{equation}

The Basic Iterative Method (BIM)\cite{kurakin2018adversarial} allows the iteration of the FGSM where the adverse effect can be increased without affecting the visual quality as much as by increasing the $\varepsilon$ by a comparable amount. BIM generates an adversarial example $i$ by applying the FGSM method to the adversarial example $i-1$. This process is laid out in the following function where $\Psi$ is the FGSM function:

\begin{equation}
    \begin{split}
        x'_0&=x \\
        x'_{i+1}&=x'_i+\Psi(\theta,x'_i,y)
    \end{split}
\end{equation}

{\color{black} The main danger of an adversarial attack on a UAV guidance system is the risk of collision with the environment it is working in. There is a secondary risk with the denial of access through guiding the UAV from the selected goal by implementing adversarial attacks. In this paper,, the adversarial attack's ability to limit the drone's successful operation will be analysed.

The adversarial attack will act upon the depth image input to the DRL agent. This will be a white-box attack as the FGSM will have access to the model to calculate the gradients to create the perturbation. In this paper, we are more interested in the detector's performance, so the attack's white-box nature is less important to the study. Attacking one of the detectors with a black-box attack can be done in future work. 

To test the capability of adversarial attack, the drone will be attacked using both the FGSM and BIM first to discover the best variations of the iteration value and the $\varepsilon$ to create the highest successful attacks while taking into account the computation time to craft the attack and the amount of distortion applied to the image. Once an attack strength is decided upon, the drone will run through the obstacle course before being attacked for 5-time steps consecutively. When the attack takes, place will be random to remove any bias. The continuous attack is chosen as one wrong move caused by an attack is unlikely to lead to a collision.}

\subsubsection{Convolutional Neural Network based Adversarial Detection}
It is proposed that detecting the slight changes in the distribution of the SHAP values from an adversarial attack is best done by a DL network. Detecting adversarial attacks in classification tasks using SHAP values is proposed by Fidel et al.(2020)\cite{fidel2020explainability}. Their classification task used a simple, Fully Connected Network (FCN) to detect adversarial examples. In our work, the task is entirely different as it is about DRL-based drone guidance tasks and a simple FCN is found to be incapable of detecting the adversarial attack. Furthermore, exploiting and adopting the explainability SHAP values for our guidance task differs from any simple classification task. Adopting an FCN-based Adversarial Detector FCN-AD in a preliminary test achieves an accuracy of only 56\%. Random chance would achieve a 50\% accuracy. Thus, a different solution is needed.

The adversarial attack only affects the depth image inputs into the DDPG-APF agent. Hence, the SHAP values for the bearing and distance to the goal will not be used for adversarial detection. The SHAP values for the depth image provide a value for each pixel, and by normalising these values, they can be turned into a new SHAP image. The normalisation can be done by either normalising the values to between 1 and 0 or by designating positive and negative to different colours and then normalising. The second option is chosen to contrast the positive and negative values. A CNN-based Adversarial Detector is proposed to take the SHAP values and decide on the input's validity.

{\color{black}CNNs have been the most used NN type for image processing tasks because of their excellent ability to detect spatial features. As the adversarial attacks only affect the depth image input to the model, it would be wise to utilise the strength of CNN in image processing to build an adversarial detector. Without using the SHAP values, the changes in the depth image would be too small for the CNN to recognise. By using the SHAP values, the change between the regular and adversarial images will be much more pronounced as different pixels are activated, creating a different pattern that can be detected. CNNs are built with two sections: the convolutional layers produce an activation map by applying filters that learn to find the features in the image. The features found in the image are passed onto the second section, which consists of the fully connected layers. Using the input of 5 time-distributed SHAP images generated from the depth images, the CNN-AD is constructed with convolutional layers that are also time-distributed. The output is a confidence score on the input being an adversarial example. The model consists of a batch normalisation layer followed by 6 time-distributed convolutional layers with a max pooling layer between every second convolutional layer. Following these layers, 5 dense layers produce the confidence score for the attack detection. The first four dense layers have a ReLU activation, while the final layer has a sigmoid activation.}

This CNN-AD method is proposed to work by the SHAP values having higher magnitudes around the edges of obstructing obstacles. Thus, a typical example perceived by the drone should show nice edges. In an adversarial attack, the importance of the real features, such as the edges, will decrease, while the pixels made more prominent by the noise generated by the FGSM attack will have higher SHAP values. 

\subsubsection{Long Short Term Memory Network-based Detection}
\begin{figure*}[htp]
    \centering
    \includegraphics[width=14cm]{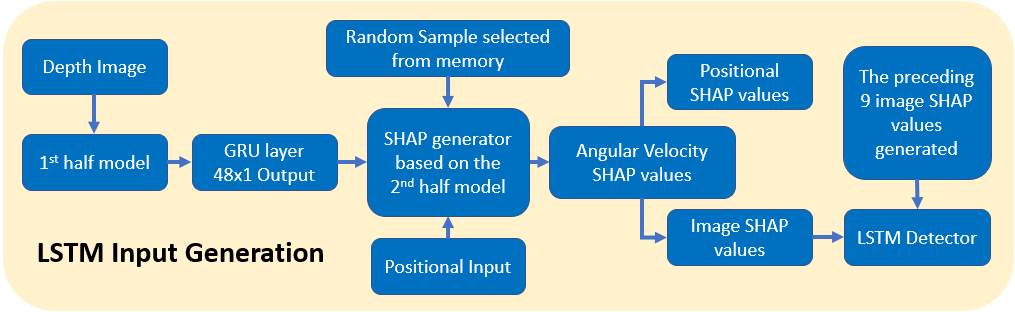}
    \caption{The information flow from the data gathered by the UAV to the detection by the LSTM-AD.}
    \label{fig:SHAP generation for LSTM}
\end{figure*}


\begin{figure*}
    \centering
    \includegraphics[width=15cm, height=6cm]{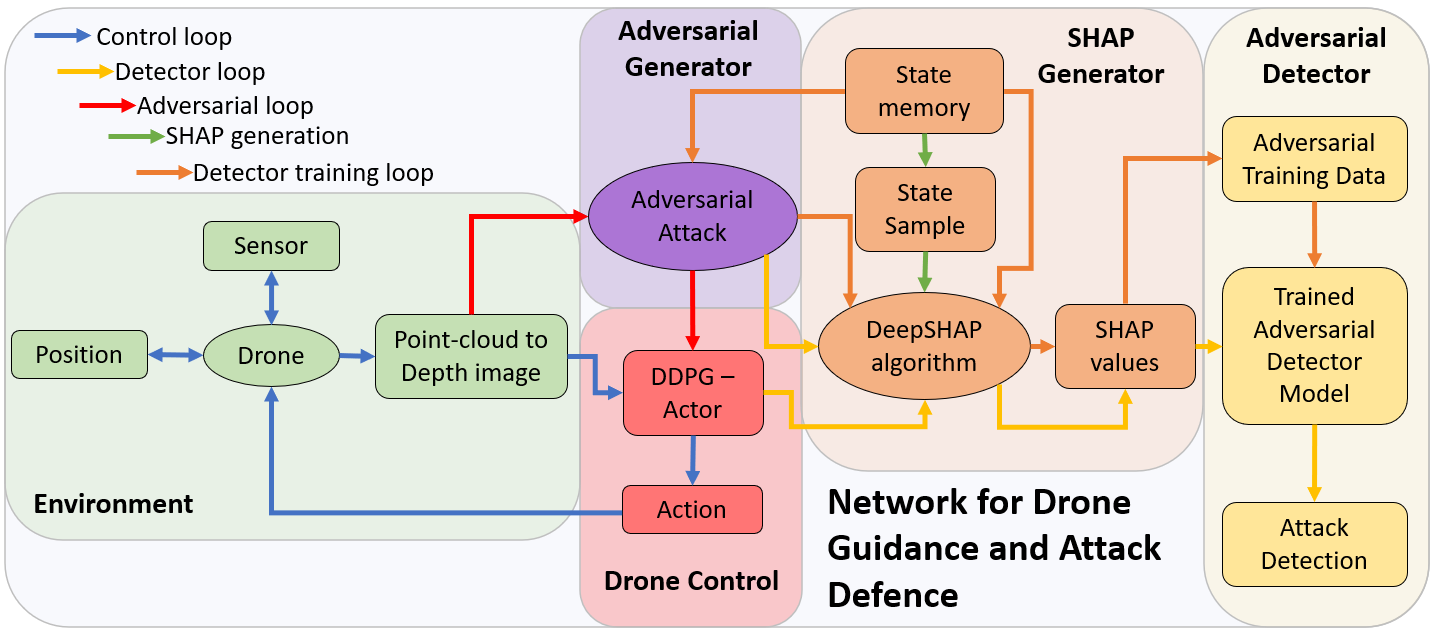}
    \caption{The layout of the experiment with the drone being controlled by the DDPG-APF network and the adversarial creation and detection using SHAP values.}
    \label{fig:project flow}
\end{figure*}

The CNN-AD presents the drawback of generating over 300,000 SHAP values every time step. Calculating a fraction of the SHAP values using a layer with fewer outputs in the DDPG-APF agent network is computationally more efficient. The smallest layer that only depends upon the depth image input is the GRU layer (see Fig. 1) with 48 values. Calculating the SHAP values for GRU layer output will be faster than for the CNN-AD method. The CNN-AD detects features in the depth image to make decisions; thus, using the 48-value vector rules out this approach. An individual 48-value vector carries a limited amount of information. The change in SHAP values can be monitored to increase the information. This increases the information available for analysis without increasing the number of SHAP values calculated each time step. 

{\color{black}RNNs are neural networks that excel in modelling temporal dependencies. It is theorised that the sudden change in how decisions are made from the effect of an adversarial attack will show up by monitoring the change in the SHAP values. So by using an RNN, the trend in the SHAP values will be detected. LSTMS are developed from RNNs to give a long-term memory as this helps decode long-term dependencies. Unlike simple RNNs, LSTMs maintain a cell state carried through the modules. This cell state is what allows the LSTM to handle long-term dependencies. LSTM networks\cite{hochreiter1997long} and are found to outperform simple RNNs\cite{prabowo2018lstm}. Our proposed LSTM-based Adversarial Detector (LSTM-AD) is looking for adversarial examples that cause enough change in the data to be detected as different to the previous time steps.} 

A smaller section of data needs to be assessed to improve the processing time for the SHAP values and limit the size of the LSTM-AD network. It is decided to measure the SHAP values at the GRU layer found after the convolutional layers in the network; this will reduce the number of values calculated from 163,840 to 48 per state. This layer's output is chosen as it is the smallest state available, only affected by the depth image input. As the adversarial attack only alters the depth image, it makes sense to consider an output only changed by the attack.

To access the data coming out of the GRU layer, the model is split into two separate models; one is just the image processing layers of the original network that outputs the 48-value state after the GRU layer; this will be known as the 1\textsuperscript{st} half model. The second is the rest of the network, which takes the standard positional input as well as the output of the GRU layer as its second input; this will be known as the 2\textsuperscript{nd} half model. Using the values generated by the 1\textsuperscript{st} half model, the 2\textsuperscript{nd} half model can generate the SHAP values required for analysis by the LSTM-AD. These SHAP values can then be stacked in 10 time-step packets for detection in the LSTM-AD. This process from depth image to inputting the stacked SHAP values into the LSTM-AD is shown in Fig. 6.{\color{black} The LSTM-AD is formed by a LSTM layer that is then followed by 4 dense layers to output a decision. The LSTM layer has 100 units and a ReLU activation; the first 3 dense layers have a ReLU activation, with the final layer having a sigmoid activation to produce the confidence score for the attack detection.}

\begin{algorithm}
{\color{black}
    \caption{Adversarial Detection}
    \begin{algorithmic}
        \State Initialise the environment $E$ and drone $A$
        \State Initialise the DeepSHAP using sample $S\subseteq F$
        \State Load Detector $D$
        \While{$A\neq Done$}
        \State Get state $s_t$ from the environment
        \If{Attacked}
            \State Set target $a'$, iteration $i$, epsilon $\varepsilon$
            \For{i}
            \State Find perturbation: 
            \State $\Psi(\theta,s_i,y)=\varepsilon.sign(\nabla_xL(\theta,s_i,y))$
            \State Modify the state $s'_{i+1}=s'_i+\Psi(\theta,s'_i,y)$
        \EndFor
        \EndIf
        \State Calculate SHAP values $\phi(A,s_t)$
        \If{D==CNN-AD}
            \State Normalise and set $\phi<0 [0,1] \Longrightarrow \Phi_{Red Channel}$
            \State Normalise and set $\phi>0 [0,1] \Longrightarrow \Phi_{Green Channel}$
        \EndIf
        \If{D==LSTM-AD}
            \State Assemble $\Phi= \begin{bmatrix} \phi_{t-9} \\ \phi_{t-8} \\ \vdots \\ \phi_t
            \end{bmatrix}$
        \EndIf
        \State Pass $D(\Phi)=[0,1]$
        \If{$D>0.5$}
            \State Attack detected stop drone
        \Else
            \State Continue
            \State Check $A = Done$
        \EndIf
        \EndWhile
        
    \end{algorithmic}
    }
\end{algorithm}

\section{Experiments and Results}\label{sec4}
The complete layout of the various subsections and how they interact can be seen in Fig. 7. The flow of information from the drone to the drone control system can be seen, and how that can be hijacked to insert adversarial attacks. The method of SHAP generation can be seen, and the use of these SHAP values to detect any adversarial attacks. 

\subsection{Experimental Setup}
To handle the simulation for this work, AirSim\cite{shah2018airsim} is used in conjunction with Epic's Unreal Engine 4 (UE4). AirSim is a high-fidelity physical and visual computer simulation that allows the generation of large amounts of training data for machine learning projects.

The test arena is set up as shown in Fig. 8. Two-goal areas are 20 metres downrange of the start position and 5 metres offset from the centre line. The goal is chosen randomly at the start of each episode to avoid the DRL-based guidance agent becoming overfitted with one solution.

\begin{figure}[b]
    \centering
    \includegraphics[width=4cm, height=4cm]{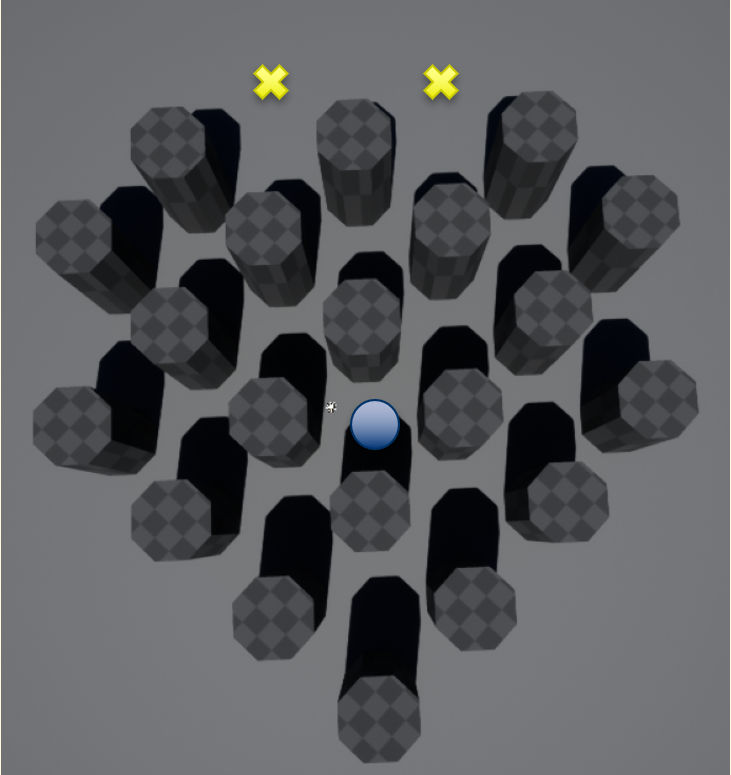}
    \caption{This figure shows the arena layout for the training of the DDPG-APF agent, with a blue circle showing the start position and yellow crosses showing the goal positions.}
    \label{fig:test layout}
\end{figure}

The simulated drone is a quadcopter. To ease training, the drone is placed into yaw mode, where it can only move forward (x-velocity, $v_x$) and change its heading (yaw rate, $\omega$). The drone is reset back to its starting position if it collides, goes out of bounds or reaches the goal.

The LiDAR sensor is excellent at quickly and precisely building a 3D environment representation. This accurate data allows for navigation in crowded environments. The data generated from LiDAR scans consist of a list of 3-D coordinates of the surrounding scene where the sensor is the origin point.

To use this data as the input into the DDPG-APF actor-network, the 3-D point cloud must be converted into a 2-D depth image to work with the convolutional layers inherited by our choice of network. Each point from the point cloud has a distance from the drone calculated, and then each value is normalised with values at 0m being 0 and values at 100m being 1; 100 m was chosen as the upper value as this was above any distance that could be measured in the simulation. These points are then projected from the 3-D point cloud to a 2-D image using the z and y components of the point cloud. Once a point is placed on the image, the pixel value is assigned as the normalised distance previously calculated. For any gaps in the created depth image, interpolation is used.

{\color{black}
The Neural Network has two outputs, one that controls the x-velocity and one that controls the yaw-rate. These two outputs are limited to between -1.5 and 1.5. These values are then handed over to a separate piece of code that broadcasts the commands to the AirSim simulation. For the x-velocity, the output from the model is directly translated into a velocity; for the yaw rate the value is multiplied by 30 to achieve a yaw rate that ranges from -45\degree/s to 45\degree/s.
}


\subsection{Reward Setup}
To train a DRL agent, getting the reward function right is imperative. The reward function is designed to  reward the drone when it moves towards the goal by adding the change in distance between the drone and the goal. There are also sub-goals that if the drone moves past a particular checkpoint, it will gain a bonus reward. The proposed reward function structure is as follows:
\begin{equation}
    R(s_t,s_{t-1})=\left \{ \begin{array}{rcl}
       5  & \mbox{if success}  \\
       -2 & \mbox{if dead} \\
       2(1+\frac{C}{N_C}) & \mbox{at checkpoint}\\
       d_t-d_{t-1}  & \mbox{otherwise}
    \end{array}\right\} = r_t
\end{equation}

Here $C$ is the checkpoint number, and $N_C$ is the number of checkpoints.

\subsection{Results}

\subsubsection{DRL Training}
The training for the DDPG-APF agent and the DDPG agent is completed successfully, and they are both able to complete the course to a satisfactory level. The training times for each model can be found in TABLE I. The two different DDPG agents' training efficiency is also compared in TABLE I, where the increase in training efficiency for the DDPG-APF agent can be seen by a decrease of 14.3\% in training time. Fig. 9 \& 10 show the progression of the reward gained and the success rate through training the DDPG-APF agent. The first 5000 time steps are to build the memory for the PER. Hence they are not included in the data as no training is done. A validation course is created for testing where the drone reached a completion rate of 80\% without APF. When the APF is applied to the layout of the validation course, the completion rises to 97\%. The run time for the DDPG-APF agent to make a decision is shown in TABLE IV, which takes around 0.05s. Overall the inclusion of the APF with the DDPG has had a negligible effect on the training efficiency though this slight increase did save a few hours as the training time was so long. The real benefit is the increase in course completion, as this added safety layer makes the agent more likely to avoid obstacles. To deploy the agent in an unknown area, a vision system must automatically place the repulsive fields to the obstacles to take advantage of the APF method.

\begin{figure}[!tbp]
  \centering
  \begin{minipage}[b]{0.49\linewidth}
    \includegraphics[width=\textwidth]{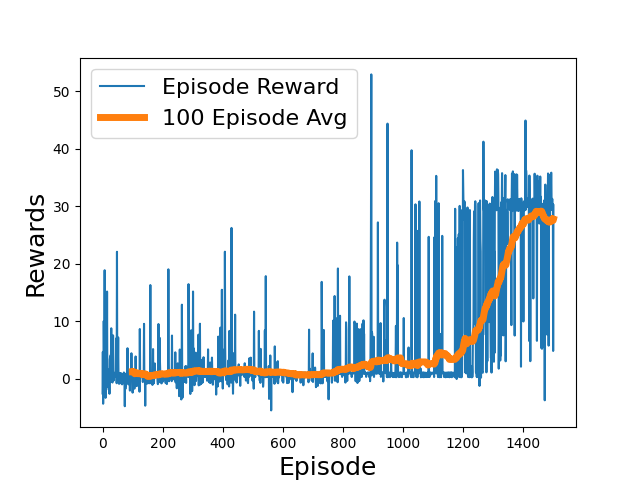}
    \caption{Reward in each episode and the rolling average.}
  \end{minipage}
  \hfill
  \begin{minipage}[b]{0.49\linewidth}
    \includegraphics[width=\textwidth]{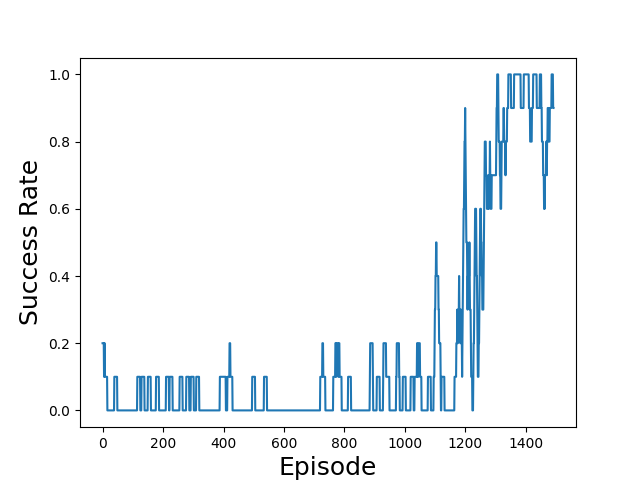}
    \caption{The success rate over the last ten episodes.}
  \end{minipage}
\end{figure} 

\begin{table}[ht!]
    \caption{Training data for the DDPG and DDPG-APF}
    \centering
    \begin{tabular}{|c|c|c|}
    \hline
         &DDPG &DDPG-APF  \\
    \hline
    Training steps & 27456 & 24012 \\
    Training episodes & 1750 & 1500 \\
    Efficiency increase & - & 14.3\% \\
    Testing success & 80\% & 97\% \\
    \hline
    \end{tabular}
    
    \label{tab:APF efficiency}
\end{table}

\subsubsection{Adversarial Examples}
The FGSM adversarial generation algorithm is implemented to easily be altered into the BIM algorithm just by running the FGSM in a loop. The FGSM is less effective than required at changing the decisions of the DDPG-APF agent at small $\varepsilon$. The BIM variation of FGSM is adopted to limit visual degradation and maximise the adversarial effect. 
\begin{figure}[htp]
    \centering
    \includegraphics[width=7cm]{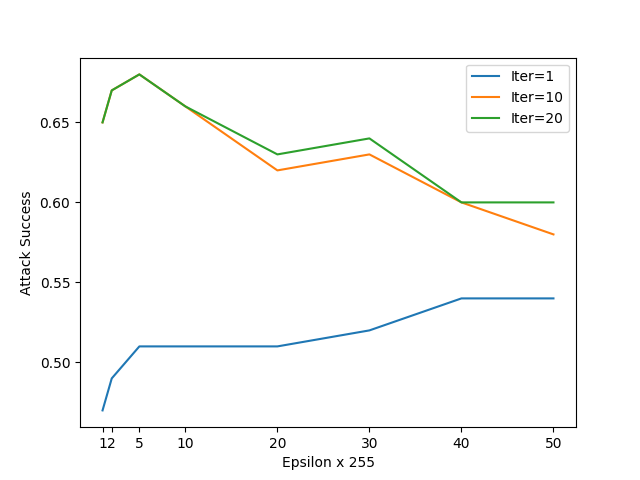}
    \caption{This figure shows that the $\varepsilon$ and iterations increase the attack success but do reach a limit in success (61.1\%).}
    \label{fig:adversarial attack}
\end{figure}

{\color{black}
We test the performance of the FGSM/BIM adversarial generator at various settings of $\varepsilon$ and the number of iterations of the FGSM for the BIM application, as shown in Fig. 11. As mentioned in the experimental setup the output of the model is a range between -1.5 and 1.5, and then this output is then multiplied by 30 to calculate a yaw-rate. To ensure an attack is significant a large deviation is chosen to be the limit for an attack to be considered successful. The guidance decision made by the DRL agent after the attack should differ by 1 from the original output or 33\% of the range. This change in the output corresponds to a change in the drone's yaw-rate of 30\degree/s. This condition on the amount of deviation ends up reaching a limit at 68\% attack success.

In Fig. 11 we can see the better performance from the BIM when compared to the FGSM. The increase in iterations from 1 to 10 or 20 improves the success rate by 15-18\%. It can also be seen that the performance peaks at an $\varepsilon$ of 5. An $\varepsilon$ with a value of 5 will produce a noticeable distortion to a depth image when compared to an $\varepsilon$ of 1 that has a distortion that is almost imperceptible. An $\varepsilon$ of 2 will have some distortion only on close inspection of the depth images. It can also be seen that the performance of the attack falls once the larger $\varepsilon$ values cause much greater distortion to the depth images.

The behaviour of the FGSM/BIM attack success in Fig. 11 requires explanation as the attack success rate falls below what would normally be seen in a classification task. The FGSM/BIM is not a perfect algorithm so failures in achieving successful attacks are to be expected, especially for drone guidance problems. A higher success rate could have been achieved by lowering the deviation required for success. Lowering this value would increase the attack success rate but it was decided only to measure large deviations that would decrease the safety of the drone the most by increasing the risk of collisions.

While adversarial attacks on classification models can perform much closer to a 100\% success rate, the model used in this situation being a DRL model is a significant difference. A classification model will degrade until a point where there are no high-confidence decisions in the classification. The DRL model will still choose a guidance output no matter how degraded the input image gets. As described in the setup for this test the range of outputs from the neural network is between -1.5 and 1.5, with a deviation of less than $\pm1$ being the limit of a measured failed attack this covers 33-66\% of the total range of actions, depending on the output before the attack. The larger distortions caused by the larger $\varepsilon$ values to the depth images mean less useful information is carried to the model and it becomes probabilistic that some attacked model inputs will fall inside the failed attack range. The problem comes when deciphering between an attack that has a negligible effect or an attack that is by chance near the original model outcome.

There are a few more explanations for why some attacked model inputs are more resistant to deviation. First, there are two different inputs for the DRL model, the depth image input as well as the positional inputs of the distance to the goal and the bearing to the goal. This paper only modelled an attack on the depth image input to the DRL model. This one input attack means this second input can override some of the attack's effects; for example, a model input that has a large bearing input will be resistant to an attack that will only increase this bearing due to the way the model has been trained to move the bearing to zero. This behaviour can be seen in Fig. 13 where the SHAP values show the impact of the bearing upon the model. Second, some model inputs are particularly resistant to attack; in the paper, Table II shows that attacks where the drone can only see the object in front of it are immune. This immunity may be because the CNN has no edges to detect, and the FGSM/BIM algorithm cannot use the edges to form an attack. Other combinations of positions and depth images resist attack either close to the goal or near the start position. 

There are limits to this analysis shown in Fig. 11, as passing or failing the attack is a binary choice that does not reflect well the magnitude of those passes or failures. A pass that deflects the Drone by the full limit of its range of motion will count as much as a deflection of the 33\% limit. 
}

Choosing the settings for the adversarial attacks depends on what factors are essential. Increasing the $\varepsilon$ increases the visual distortion of the depth input image. Increasing the perturbation by too much means the attack can be detected by close visual inspection. Increasing the number of iterations increases the strength of the attack but at a lesser cost in visual distortion when compared to just increasing the $\varepsilon$. This also increases, but moderately, the computing time by the number of iterations.

To further test the ability of the FGSM/BIM adversarial algorithm when near an obstacle. To test this, the drone is placed at various distances (1m, 1.5m, 2m, 2.5m, 3m) to the obstacle and bearings (-30\degree, -20\degree, -10\degree, 0\degree, 10\degree, 20\degree, 30\degree) to the goal. The difference in the yaw velocity decision between the adversarial example and the unaltered example is noted in TABLE II.

\begin{table}[ht!]
    \caption{Deflection caused by the adversarial attack. $\varepsilon$ of 1 and iteration of 20.}
    \centering
    \begin{tabular}{|p{1cm}|p{0.65cm}|p{0.65cm}|p{0.65cm}|p{0.5cm}|p{0.5cm}|p{0.5cm}|p{0.5cm}|}
    \hline
  Gap & \multicolumn{7}{|c|}{Bearing to goal (Degrees)} \\
    \cline{2-8}
    to wall & -30\degree & -20\degree & -10\degree & 0\degree & 10\degree & 20\degree & 30\degree \\
         \cline{2-8}
  (m) & \multicolumn{7}{|c|}{Change in yaw velocity from attack (\degree/s)}\\
         \hline
        1.0 & 0 & 0 & 0 & 0 & 0 & 0 & 0 \\
        1.5 & 0 & 0 & 61.5 & 62.1 & 62.4 & 62.7 & 63.0 \\
        2.0 & 52.8 & 53.4 & 53.4 & 53.4 & 53.1 & 53.1 & 53.1 \\
        2.5 & 30.3 & 28.8 & 26.7 & 30.6 & 32.1 & 30.9 & 32.4 \\
        3.0 & 24.9 & 24.6 & 21.0 & 24.6 & 24.9 & 25.8 & 26.4 \\
        \hline
    \end{tabular}
    
    \label{tab:column attack}
\end{table}

In TABLE II, the difference between the two decisions becomes more significant as the drone approaches the column. This difference increases as the drone produces more drastic actions to avoid a collision with the obstacle. This gives the adversarial attack a more significant decision to alter. The drone could not be attacked when very close to the obstacle. It appears that the small perturbation cannot overcome the depth image being filled with the obstacle at close range. 

The results shown are with a $\varepsilon$ of 1 and an iteration of 20. Attacks at various iterations were carried out. At five iterations, all the variations that aren't 0 in the table were over the 33\% barrier defined earlier. No matter the attack strength, the 0 values from the table remained 0.

A single attack on the drone is not a severe threat to the safety of the drone, as it will recover and continue towards its goal after the attack. A constant attack over a series of time steps must be executed to increase the danger to the drone. To test the success of this approach, the drone is set up to run through the validation course, and attacks are set up to occur randomly and run for 5-time steps. Doing this leads to completing the course 35\% of the time. This is a drop of 62\% from what the drone achieved when the drone was not attacked. The APF was not activated to simulate the agent's inability to identify obstacles when attacked. A 100\% failure rate is unlikely due to the intrinsic robustness nature of the DDPG-APF agent and the chance of the attacks coming at the wrong times.

\subsubsection{Shapley Values}
The SHAP values are first produced to analyse the inputs of the whole DDPG-APF model. It is one value for every pixel in the 5-time step depth images that make up the time-distributed depth image input and the ones for positional inputs (bearing, distance to goal). SHAP values are a measure of how much each input has an impact on the output. As the DDPG-APF model has two outputs (x-velocity and yaw rate), each decision has a set of SHAP values. For this study, the SHAP values relating to the yaw rate decision are chosen because this set of SHAP values has the most variability throughout a guidance task. The x-velocity remains stable at its maximum throughout each guidance task. Thus, there is only a slight variation in SHAP values to analyse. 

Fig. 12 shows a typical SHAP value representation of the response to the depth image. The red positive SHAP values of the right side of the column show a largely positive response favouring a right turn to avoid the collision. A sizeable negative response can be seen on the left side of the column where the SHAP values support a left turn to avoid a collision. 

\begin{figure}[htp]
    \centering
    \includegraphics[width=7.5cm]{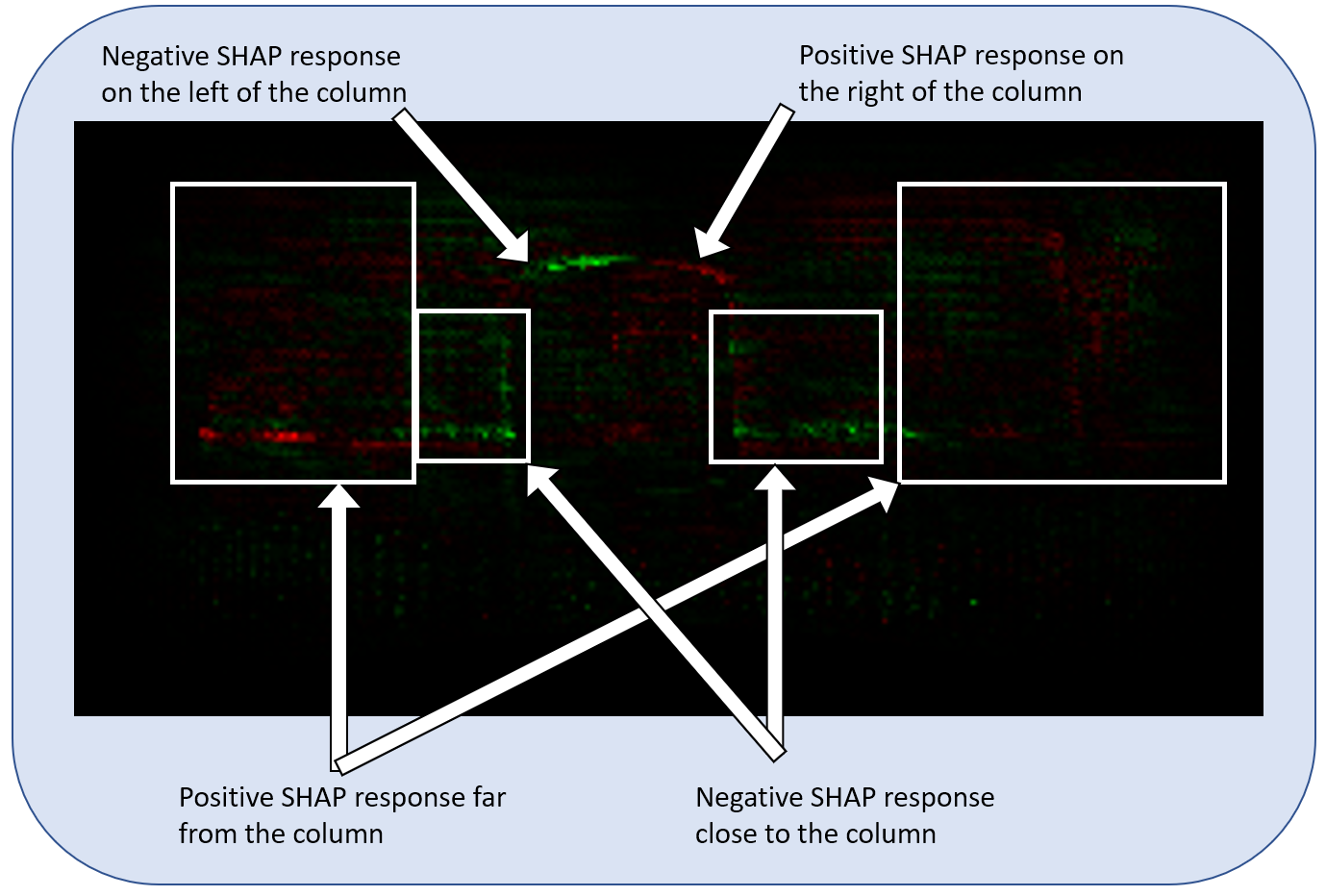}
    \caption{The figure shows SHAP values in response to a depth image. Green is a negative SHAP value, and red is a positive one.}
    \label{fig: SHAP CNN image}
\end{figure}

Though producing SHAP values for individual pixels can help visually inspect what the DDPG-APF model thinks is important, it does not allow an easy deduction of the action suggested. Thus a more parsable form of the SHAP values is created. As described in the final part of the LSTM network detection section a break is placed in the actor-network model to allow the SHAP value generation at the layer of the break. This gives 48 SHAP values which is a more manageable number to manipulate. 

{\color{black}
To analyse how the decisions made by the DDPG-APF agent are influenced by the three inputs (depth image, bearing and distance to goal), Fig. 13 is produced. The variation of the SHAP values related to the yaw rate decision over one navigation of the validation course will be plotted. The SHAP values for the positional inputs are plotted directly. For the contribution of the depth image, the 48 values are summed to give one SHAP value that represents the collective decision for that input is plotted. In Fig. 14, the yaw rate is plotted; this value should summate the SHAP values from the combined depth image, bearing and distance to the goal. The sum of the SHAP values is also plotted on this graph to validate this.

In Fig. 13, the distance to goal SHAP value has little effect on the turning decision from the DDPG-APF due to this input having no information relevant to this decision. The bearing SHAP value tracks the bearing directly by trying to turn the drone back to a bearing of 0 at all times. The depth image SHAP value responds to the obstacles in the drone's path. When the drone is forced from the 0\degree bearing by an obstacle, the bearing SHAP value will increase in magnitude to turn the drone back, but the depth image SHAP value will stay high to keep the drone from turning back into the obstacle. This behaviour is shown in Fig. 13 between time steps 25 and 33. In Fig. 14, the sum of the SHAP values is almost the same as the decision made by the model. There is a slight difference that is constant across the whole run. This value describes the natural tendency for the model to turn one way without outside input.

\begin{figure}[htp]
    \centering
    \includegraphics[width=7.5cm]{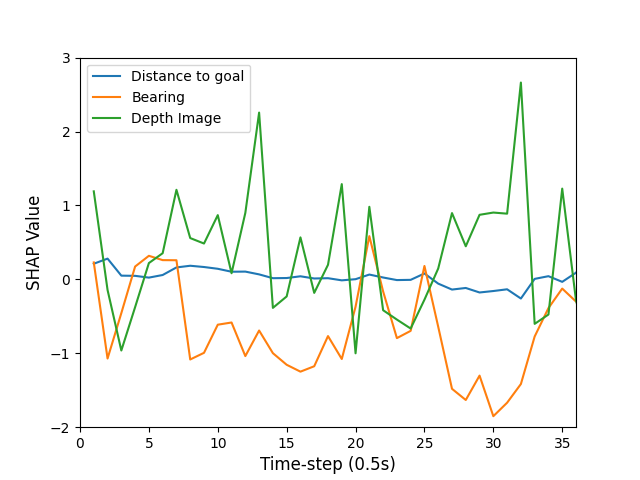}
    \caption{This figure shows the SHAP values produced by the three major inputs to the DDPG-APF model on a run through the validation course.}
    \label{fig:SHAP response}
\end{figure}
\begin{figure}[htp]
    \centering
    \includegraphics[width=7.5cm]{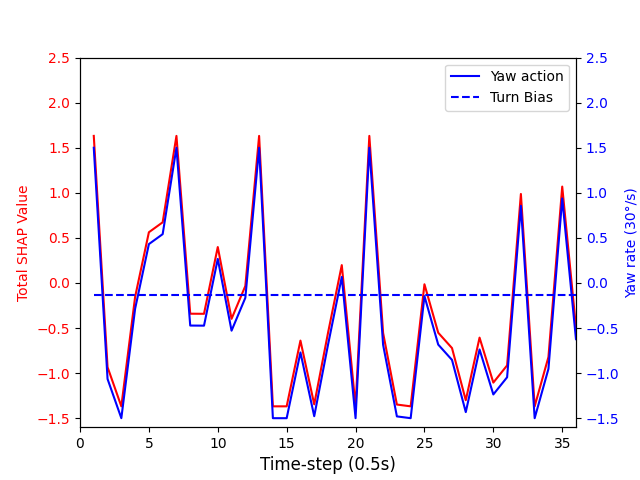}
    \caption{This figure shows the total SHAP value produced for the yaw action for the DDPG-APF model on a run through the validation course. This is shown against the actual yaw action chosen. The difference in the two lines shows that the model is biased to turn slightly to the left.}
    \label{fig:SHAP vs action}
\end{figure}}

\subsubsection{Convolutional Neural Network based Adversarial Detector}
{\color{black}
The training of the CNN-AD uses 4000 states from those generated during the training. An adversarial example is created using an iteration of 20 and an $\varepsilon$ of 1. With these states, SHAP values are created using a sample of 30 states to build the SHAP generator. The detector is then trained for ten epochs using the SGD optimiser, reaching an accuracy of 97\% in the training data and a validation accuracy of 90\%. The CNN-AD achieves an accurate detection rate of 80\% when tested across 1000 states encountered in simulated runs through the validation course.

\begin{figure}
    \centering
    \includegraphics[width=7.5cm]{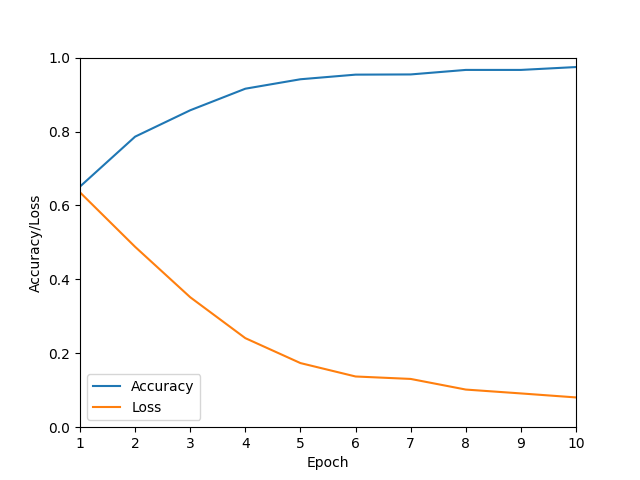}
    \caption{The training for the CNN-AD can be seen in this figure. The detector was trained over 10 epochs with an SGD optimiser.}
    \label{fig:my_label}
\end{figure}
}
The testing is done on the validation course. The drone then runs through the course with a normal and an adversarial state generated at each time step. SHAP values are then generated and then put through the detector. The detector then puts a value between 1 and 0 for each example state; a value over 0.5 is considered an adversarial detection. The closer to 1 as a score is, the more confident that it is an adversarial example. The accuracy of this detector is then measured, and if it is a wrong classification, it is assigned to being a false positive or a false negative. The testing runs for 1000 time steps and produces 1000 unaltered and 1000 adversarial states. The accuracy is the number of correctly assigned states divided by the total number of states.  

The accuracy of the CNN-AD is 80\% which leaves 20\% either undetected or wrongly classified as an attack. A hypothesis that higher accuracy is not achievable in this setup is that the ability of the detector to distinguish types depends upon the adversarial attack creating a different outcome. As shown in the previous section, large deviations only appear in 61\% of cases. It is reasonable to assume that with small or non-existent deviations, the change in the SHAP values structure might not be detectable for this CNN-AD. 

An issue with this method is the time consumed to create the SHAP values for the detector to use. The SHAP value generation takes between 20.8s and 21.1s at each time step in its current setup. The current 21s would mean that only one sample could be done over the drone's course navigation, which is impractical.

\subsubsection{Long Short Term Memory based Adversarial Detector}
{\color{black}
Building upon the success of the CNN-AD, an LSTM network is created to attempt to detect adversarial examples. The concept here is to detect an anomaly in the signal generated over 10-time steps of the SHAP values of the GRU layer, shown in Fig. 1, as described in earlier sections. The flow of data generation for the LSTM-AD is demonstrated in Fig. 6. The method to generate the training samples for the CNN-AD does not work for the LSTM-AD detector, as combining states blindly can lead to poor examples. A different approach is proposed by creating samples directly from the simulation. This is done by running the drone through the training course generating an adversarial and normal example at each time step. This method gathered 15,000 samples of each type for the training. The training was then run over 1000 epochs with an AdaDelta optimiser.

\begin{figure}
    \centering
    \includegraphics[width=7.5cm]{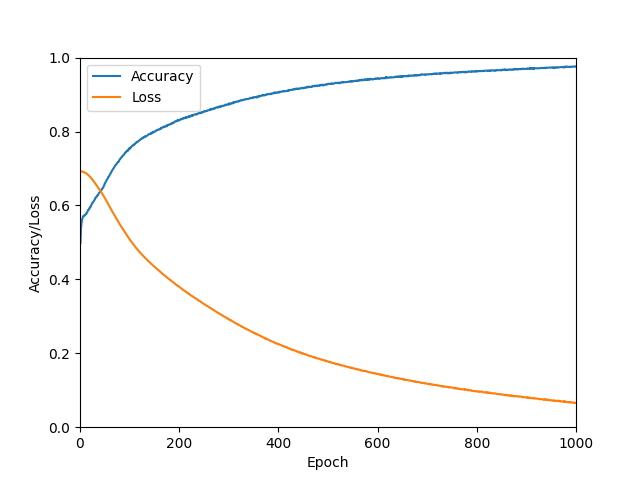}
    \caption{This shows the progression of the accuracy and loss of the LSTM-AD model over the 1000 epochs of training with an AdaDelta optimiser.}
    \label{fig:LSTM_training}
\end{figure}
}
The LSTM-AD's training parameters are different to those of the CNN-AD. Many more epochs are used for the LSTM-AD; the optimal is found at 1000 epochs. The accuracy for the training data reaches 100\%, but to get the best accuracy in the validation, the number of epochs is limited to the 1000 epochs mentioned above, which gives a training accuracy of 95\% and testing at 90\%. The validation accuracy at 91\% was much closer to the testing accuracy with this network.

When compared to the CNN-AD, this method is faster to compute. In TABLE IV, a reduction of 99.9\% from 13.33s to 0.019s for calculating the SHAP values is observed. When looking at the adversarial detectors' compute time, a similar decrease of 99.4\% from 0.162s to 0.001s is observed. This enables the detector to analyse every time step for potential attacks. This method also led to an increase in accuracy over the CNN-AD by 10\%. This increase shows that the LSTM-AD is better at identifying the differences the CNN-AD  could not detect.

{\color{black}
Comparing the results of these two adversarial detection methods to other methods is difficult due to the lack of studies on adversarial detection for drone guidance and navigation tasks. Most studies in this area work by making the agent robust so that they are not affected by adversarial attacks, but this doesn't inform if the agent is being attacked. It is, therefore, difficult to compare the proposed methods where we are interested in if an attack has taken place. Due to the lack of comparisons from drone guidance studies from other areas that are to be used, these may not give a one-to-one comparison. Tekgul et al.(2022)\cite{tekgul2022real} only give detections over a whole episode by either looking at a trend through the episode or by having multiple chances at detection. These methods provide higher detection rates than the spot detections done in this paper. Other papers, such as Lin et al.(2017)\cite{lin2017detecting}, remove the unsuccessful attacks from their detection data leading to a much higher detection rate of over 99\%.

In TABLE III, we can compare the methods detailed in this paper and a comparative method using Principle Component Analysis (PCA)\cite{xiang2018pca}. The report looks at a 2D map navigation task, so this is not a direct comparison between their work and this paper. Even so, their detection rate was at 70\%; this is a lower performance than the CNN-AD and LSTM-AD built for this study. Compared to all adversarial detection schemes the LSTM-AD is the only one that can claim to be a real-time detection method.
}
\begin{table}[ht!]
    \caption{Comparison of the detection methods to other studies.}
    \centering
    \begin{tabular}{|p{1.4cm}|p{1cm}|p{1.5cm}|p{2.3cm}|}
    \hline
    & \multicolumn{3}{|c|}{Run time for each section (s)} \\
    \hline
    Detector type & Detection Rate & Agent Model  & Task \\
    \hline
    FCN-AD & 56 & DDPG & Drone Guidance \\
    CNN-AD & 80 & DDPG & Drone Guidance \\
    LSTM-AD & 91 & DDPG & Drone Guidance \\
    PCA\cite{xiang2018pca} & 70 & Q-learning & Map Navigation \\
    \hline
    \end{tabular}
    
    \label{tab: Comparison of Adversarial Detection Methods}
\end{table}
\begin{table}[ht!]
    \caption{Times taken to produce the required data for the DRL, XAI and Adversarial Detection sections.}
    \centering
    \begin{tabular}{|p{1.4cm}|p{1.7cm}|p{1.8cm}|p{1.7cm}|}
    \hline
    & \multicolumn{3}{|c|}{Run time for each section (s)} \\
    \hline
    Detector type & DDPG-APF Action & SHAP Values Generation  & Adversarial Detection \\
    \hline
    FCN-AD & 0.053 & 13.14 & 0.056 \\
    CNN-AD & 0.052 & 13.33 & 0.162 \\
    LSTM-AD & 0.056 & 0.019 & 0.001 \\
    \hline
    \end{tabular}
    
    \label{tab:timings of computation}
\end{table}

\section{Conclusions}\label{sec5}

This paper sets out four goals. The first is to test the performance increase from the APF addition to the DDPG model. The drone is successfully trained and achieves a good completion percentage. The adoption of APF shows the training efficiency increase of the DRL model and an increase of 17\% to validation course completion. The APF module could be added to any DRL scheme, and a future study may look into whether the improvement seen here is achieved in more DRL schemes. The second goal is to apply the DeepSHAP algorithm to this model to generate individual input-level explanations. This is completed with explanations being generated and images being made out of these explanations to show where the drone's attention is when navigating. The third goal is to successfully generate adversarial examples that can trick the drone into making poor guidance decisions. This is completed using the FGSM and BIM algorithms, which reduces guidance success by 56\%. The fourth goal is to design a method for detecting adversarial attacks by adopting SHAP values-based explanations. This is completed in two ways, first by building the CNN-AD that gained an accuracy of 80\%. The second is creating a detector based on an LSTM neural network module that looks for changes in the signal produced by adversarial attacks. The LSTM-AD delivers an accuracy of 90\% with a lower computation cost, enabling real-time detection compared to the CNN-AD. 

The lack of comparable results to those gained in this paper is due to either differing applications or different methods of analysis of the success of an adversarial detector. This has made an overall conclusion of the success of these DNN adversarial detectors in a broader context. To find which methods are the best at dealing with adversarial attacks, a more standardised set of measures of success to be measured by would be advisable.

A method to get, for future research, the correct decision after detecting an adversarial attack would close off the problem as there is no real recourse after detecting the attack at the moment. Bringing this into the real world would also be a promising avenue of study as a method for creating real-world adversarial attacks is more complicated than just generating them for a simulation. 

\bibliographystyle{plain}
\bibliography{refs}  

\end{document}